\documentclass{article}

\PassOptionsToPackage{numbers, sort&compress}{natbib}

\usepackage[final]{neurips_distshift_2023}

\usepackage[utf8]{inputenc}

\usepackage[T1]{fontenc}    %
\usepackage{url}            %
\usepackage{booktabs}       %
\usepackage{amsfonts}       %
\usepackage{nicefrac}       %
\usepackage{microtype}      %
\usepackage{amsmath, amsthm, graphicx, amssymb}
\usepackage[dvipsnames]{xcolor}
\usepackage[colorlinks,linktoc=all,pagebackref=true]{hyperref}
\hypersetup{
    linkcolor=BrickRed,
    citecolor=PineGreen,
    filecolor=Mulberry,
    urlcolor=NavyBlue,
    menucolor=BrickRed,
    runcolor=Mulberry,
}

\usepackage[noend]{algpseudocode}
\usepackage{fancyhdr}
\usepackage{algorithm,algpseudocode,tabularx}

\usepackage{multirow,bm,enumitem,xfrac,pgf}
\usepackage{amsmath,graphicx,mathtools}
\usepackage{xspace,capt-of,float}
\usepackage[title]{appendix}
\usepackage{booktabs, tabularx}
\usepackage{enumitem, wrapfig, multirow}
\usepackage[capitalise,noabbrev]{cleveref}
\usepackage{annotate-equations, circledsteps, todonotes}

\DeclareUnicodeCharacter{2212}{-}

\NewCommandCopy{\ORIcitep}{\citep}
\DeclareRobustCommand{\citep}{\leavevmode\unskip~\ORIcitep}

\NewCommandCopy{\ORIcitet}{\citet}
\DeclareRobustCommand{\citet}{\leavevmode\unskip~\ORIcitet}
\AtBeginEnvironment{appendices}{\crefalias{section}{appendix}}
\AtBeginEnvironment{appendices}{\crefalias{subsection}{appendix}}
\newcommand{\propmethod}{CoLoR}

\makeatletter
\DeclareRobustCommand\onedot{\futurelet\@let@token\@onedot}
\def\@onedot{\ifx\@let@token.\else.\null\fi\xspace}

\def\ie{\emph{i.e}\onedot,\xspace}

\makeatother
\crefformat{section}{\S#2#1#3}
\crefmultiformat{section}{\S#2#1#3}{ and~#2#1#3}{, #2#1#3}{, and~#2#1#3}

\def\W{\ensuremath{\bm{W}}}

\usepackage{array}
\newcolumntype{H}{>{\setbox0=\hbox\bgroup}c<{\egroup}@{}}

\title{Continual Learning with Low Rank Adaptation}

\author{%
    Martin Wistuba\\ 
    Amazon Web Services\\
    \And Prabhu Teja S \\
    Amazon Web Services \\\AND
    Lukas Balles\\
    Amazon Web Services  \\ 
    \And Giovanni Zappella\\
    Amazon Web Services \\
}

\begin{document}

\maketitle

\begin{abstract}
\begin{itemize}
    Recent work using pretrained transformers has shown impressive performance when fine-tuned with data from the
    downstream problem of interest.
    However, they struggle to retain that performance when the data characteristics changes.
    In this paper, we focus on continual learning, where a pre-trained transformer is updated to perform well
    on new data, while retaining its performance on data it was previously trained on. Earlier works have tackled this primarily through methods inspired from prompt tuning.
    We question this choice, and investigate the applicability of Low Rank Adaptation (LoRA) to continual learning. 
    On a range of domain-incremental learning benchmarks, our LoRA-based solution, \propmethod{}, yields state-of-the-art performance, while still being
    as parameter efficient as the prompt tuning based methods.
\end{itemize}
\end{abstract}

\section{Introduction}\label{sec:intro}
A primal feature of human cognitive abilities is to incrementally and continually update knowledge of a problem;
a child can seamlessly learn to recognize newer breeds of dogs without forgetting previously learned ones. Modern
machine learning systems, however, fail at this. When na\"ive methods for fine-tuning are used to update the weights,
they perform well on the specific dataset it has been fine-tuned on, while losing performance on previous ones, a phenomenon
called \emph{catastrophic forgetting}\citep{french1999catastrophic,mccloskey1989catastrophic}. This issue, while
not as drastic for modern pre-trained transformers\citep{ramasesh2022effect}, is still a major hindrance to 
the deployment of reliable systems. Continual learning\citep{parisi2019continual,de2021continual} deals with this 
problem of periodically updating a model with new data, while avoiding forgetting previous information. 

In practice, data arrives as a sequence of datasets and we aim at performing well on the latest dataset while retaining performance on the previous ones. Several paradigms of continual learning are defined based on the 
differences between each dataset. 
In domain-incremental learning (DIL), the set of labels is fixed, whereas the data distribution can change arbitrarily. 
In class-incremental learning (CIL), the set of labels is growing with new datasets which poses the challenge of recognizing newly introduced classes.
In task-incremental learning (TIL), we learn to solve different tasks and the number of tasks grows incrementally.
At training and prediction time, we are aware of the task identity which is not the case in the other settings.

With transformer-based models becoming commonplace, several continual learning methods have been proposed that use specific architectural components of those models. These methods
are heavily inspired by the parameter-efficient fine-tuning methods in NLP\citep{ruder-etal-2022-modular}, primarily, prompt tuning\citep{lester-etal-2021-power}. Prompt tuning prepends a set of learnable parameters to the outputs of the input embedding layer and trains only those,
while keeping the rest of the model frozen. Learning to Prompt (L2P)\citep{zhou2021learning} trains a set of input-dependent prompts that are shared across datasets, which encourages transfer. S-Prompts\citep{wang2022sprompts} instead learns a single prompt per dataset, and proposes a method to determine which prompt to use at inference. We discuss several other works in \cref{sec:prior_work}.
However, the choice of using prompt tuning is not justified sufficiently in these methods beyond parameter-efficiency, despite prior work\citep{su-etal-2022-transferability, hu2022lora} demonstrating prompt tuning is slower to train and achieves lower test time performance than the full fine-tuning counterpart. 

In this work we revisit this choice, in light of evidence from the NLP community that shows low rank update methods\citep{hu2022lora} perform better than prompt-based ones. 
We propose an adaptation of S-Prompts, the state-of-the-art for domain-incremental learning, called \propmethod{} for efficient continual training of vision transformers showing a significant improvement in predictive performance.
With an empirical evaluation on three domain-incremental benchmarks, we show that \propmethod{} outperforms prompt-based methods such as L2P and S-Prompts in terms of average accuracy and forgetting.
Furthermore, we show that these gains are achieved with approximately the same number of model parameters.
We propose a simple extension to our method called \propmethod{++} that yields state-of-the-art results on Split CIFAR-100.

\section{Continual Low Rank Adaptation}\label{sec:proposed_method}
We, discuss Low Rank Adaptation (LoRA), and then present our method Continual Low Rank Adaptation (\propmethod{}).

\subsection{Low Rank Adaptation}\label{subsec:lora}
We focus on vision transformers in this work, but this approach is sufficiently general to be used with other pre-trained transformers. A detailed description of vision transformers is provided in \cref{app:sec:vit}. 
Traditional fine-tuning updates all the weights of a pre-trained transformer with the data of a downstream task. Low Rank Adaptation\citep{hu2022lora} constraints the update to a low rank one. An update to a parameter matrix $\W \in \mathbb{R}^{d\times k}$ of the form $\W \leftarrow \W + \Delta \W$ is constrained by parameterizing $\Delta \W = \bm{B}\bm{A}$ where $\bm{A}\in\mathbb{R}^{r\times k}$ and $\bm{B}\in\mathbb{R}^{d\times r}$. This restricts $\Delta \W$ to a rank $r$, and is also parameter-efficient; when $r \ll k$, the total number of parameters that are updated is $r(d+k)$ instead of $kd$ as is in the case of full fine-tuning. In addition, LoRA is applied only to query and value embedding matrices ($\W_Q$ and $\W_V$) in all the layers of the network, thereby further reducing the number of trainable parameters compared to full fine-tuning.
At inference, the added parameters can be merged with the old parameters, keeping the inference time unaffected.

\subsection{\propmethod{} -- Training and Inference}

\paragraph{Training}
\propmethod{} leverages a pretrained model $h$ and extends it using LoRA to train an expert model for each dataset $D$.
Let us denote the expert model for dataset $D$ with $f_D(\mathbf{x})=g_D\circ h(\mathbf{x};\Theta_D)$ where the parameters of $h$ are frozen but it is extended by dataset-specific LoRA modules parameterized by $\Theta_D$. $g_D$ refers to the dataset-specific classification layer $g_D(\mathbf{x}) = \text{softmax}(\mathbf{w}_D^{\intercal} h(\mathbf{x};\Theta_D) + b_D)$ which uses the [CLS] token of the vision transformer. The trainable parameters of the network are $\Theta_{D,l}=\{\bm{A}^{t,l}_Q, \bm{B}^{D,l}_Q, \bm{A}^{D,l}_V, \bm{B}^{D,l}_V\}$ corresponding to all LoRA components added to each layer $l$, and the parameters of the classifier $\mathbf{w}_D, b_D$. The overall network is trained with a loss appropriate for the downstream problem.

\paragraph{Inference}
As the dataset identifier $D$ is not available at inference time, we use a simple unsupervised method\citep{wang2022sprompts} to infer it.
We estimate $k$ dataset prototype vectors for each dataset $D$ at training time as follows.
First, we embed each training instance using $h$ (without LoRA modules), and run $k$-means on those feature embeddings.
We store the $k$ cluster centers which serve as representatives for dataset $D$.
At inference time for an instance $\mathbf{x}$, we estimate the cluster center which is nearest to $h(\mathbf{x})$.
Then, we use $f_{\hat{D}}$ to make the prediction for $\mathbf{x}$, where $\hat{D}$ is the dataset corresponding to the nearest cluster center.

\section{Experiments}\label{sec:experiments}
\paragraph{Experimental setup}
Our experiments closely mirror those of \citep{wang2022sprompts}. For domain incremental learning experiments, we show results on CORe50\citep{lomonaco2017core50} and DomainNet\citep{peng2019moment}. CORe50 is a benchmark for continual object recognition with 50 classes from 11 datasets with 8 of them acting as the training set, and the rest as the test set. DomainNet is a benchmark for image classification with 345 classes and 6 datasets. For class incremental experiments, we use Split CIFAR-100\citep{zenke2017continual} which splits the CIFAR-100 into 10 datasets of 10 contiguous classes each.

To facilitate a fair comparison of baselines, we use a ViT-B-16 model\citep{dosovitskiy2020vit} pretrained on ImageNet21k from the timm library\citep{rw2019timm}, and report average accuracy, \ie the fraction of correctly classified test instances up to the current dataset. Our code base is built on top of S-Prompts\citep{wang2022sprompts}.

We provide a summary of our results here, and present detailed tables in \cref{app:results} (\cref{tab:dil,tab:cil,tab:dil-vs-til}). We, primarily, focus on memory-free methods here and relegate a broader comparison with replay-based methods to the Appendix. %

\begin{figure}[ht]
  \centering
  \renewcommand\sffamily{}
  \input{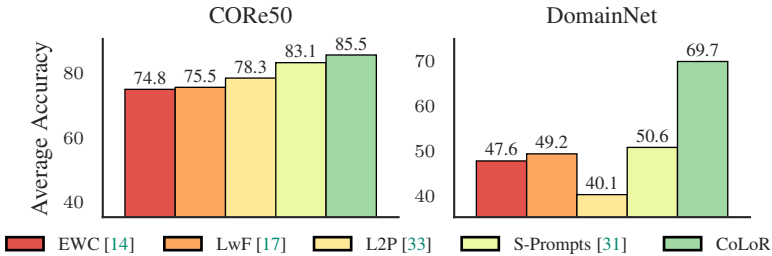}
  \caption{
    Results on two different datasets for domain-incremental learning.
    \propmethod{} improves by 2\%-19\% over the next best memory-free method.
  }
  \label{fig:dil}
\end{figure}

\paragraph{\propmethod{} demonstrates new state-of-the-art results in domain-incremental learning.}
In \cref{fig:dil}, \propmethod{} demonstrates superior performance compared to all other methods. It outperforms its closest competitor by 2\% on CORe50, and 19\% on DomainNet.
Furthermore, \propmethod{} performs on par or better than replay-based methods (Appendix, \cref{tab:dil}).

\begin{wrapfigure}[19]{r}{.48\linewidth}
  \centering
  \renewcommand\sffamily{}

  \input{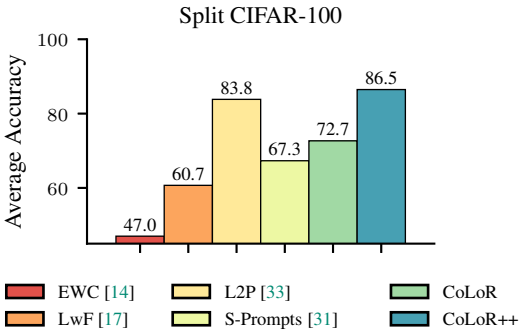}
  \caption{\propmethod{} improves by more than 5\% on CIFAR-100 in the class-incremental scenario over S-Prompts.}
  \label{fig:cil}
\end{wrapfigure}

\paragraph{LoRA is beneficial in class-incremental learning.}
Results on Split CIFAR-100 support our argument that LoRA is a better choice than prompt tuning, as \propmethod{} yields better results than S-Prompts (\cref{fig:cil}).
However, \propmethod{} lags behind L2P due to the quality of representations extracted by ViT ($h(\cdot)$) for the dataset identification method.
To address this shortcoming, we propose the \propmethod{++}, which uses the representation extracted by the network after the first dataset update, \ie $h(x, \Theta_1)$.
We believe that this feature extractor effectively represents the data as it has been trained on a portion of it, leading to improved results.
A comparable enhancement is also noticed in domain-incremental learning, albeit to a lesser extent (Appendix, \cref{tab:dil}).

\paragraph{\propmethod{} retains the parameter-efficiency of S-Prompts}
\cref{tab:dil-vs-til} summarizes the additional parameters required for \propmethod{} and its
prompt-tuning competitors on a hypothetical two class problem.
Since this efficiency holds only true for low ranks $r$, we report the additional accuracy results in \cref{fig:ablation-rank} and \cref{tab:cil,tab:dil} in the Appendix.
It is apparent, that for the same number of parameters, \propmethod{} still provides better results than its competitors.
Furthermore, increasing the rank allows to trade parameter-efficiency for prediction performance.

\paragraph{\propmethod{} closes the gap between DIL and TIL.}
In previous experiments, we assume no access to the dataset identifier at inference, and use our dataset identification method to determine which LoRA module to use.
In \cref{tab:dil-vs-til}, we show the results for using an oracle dataset identification method.
A substantial increase in accuracy is expected as the dataset identification is non-trivial; in particular, in CIL a wrongful dataset prediction leads to a mis-classification.
However, for DIL this happens to a lesser degree and \propmethod{} closes the gap between TIL and DIL.
Finally, TIL performance can be construed to be the upper bound of using LoRA-based modules for continual learning.
Importantly, this upper bound is significantly higher than the one oftentimes attained by training a single model using all data (see Appendix, \cref{tab:cil}).

\newcommand{\gray}[1]{\textcolor{gray}{\scriptsize \ensuremath{#1\uparrow}}}
\begin{table}[t]
  \begin{center}
    \caption{
      Number of trainable parameters for each method.
      We report the parameters trained for DyTox, L2P, S-Prompts, and \propmethod{} ($r=1$) for a
      hypothetical two class problem. $\dagger$ numbers are reproduced from \citep{wang2022sprompts}.
    } \label{tab:param-efficiency}
    \begin{tabular}{lcccc}
      \toprule
                                                                      & DyTox$\dagger$ & L2P$\dagger$  & S-Prompts$\dagger$ & \propmethod{} \\
      \midrule
      \multirow{2}{*}{Additional Parameters per Dataset (on average)} & 1.42M          & 18.43K        & 52.22K             & 38.40K        \\
                                                                      & \gray{1.65\%}  & \gray{0.02\%} & \gray{0.06\%}      & \gray{0.04\%} \\
      \bottomrule
    \end{tabular}
  \end{center}

\end{table}

\begin{figure}
  \centering
  \renewcommand\sffamily{}
  \input{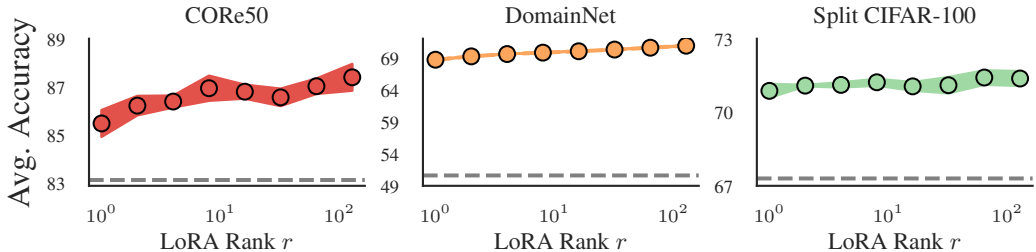}
  \caption{Increasing the rank by keeping all other settings fixed. Increasing the rank beyond 2-digit numbers yields only minor improvements in most cases.
    \propmethod{} outperforms its best competitor even with the smallest rank.}
  \label{fig:ablation-rank}
\end{figure}

\begin{table}[H]
  \caption{Inferred dataset id vs known dataset id with \propmethod{}. We report the performance in the case where the dataset id is inferred as explained above and in the case there the correct dataset id is provided by an oracle. While the oracle-based setting is not realistic, this comparison is still useful to investigate the performance of the algorithm.
    This experiment is not applicable for CORe50.
  }
  \label{tab:dil-vs-til}
  \centering
  \begin{tabular}{lHcc}
    \toprule
                                      & CDDB  & DomainNet & Split CIFAR-100 \\
    \midrule
    \propmethod~(inferred dataset id) & 85.00 & 69.67     & 71.42           \\
    \propmethod~(correct dataset id)  & 86.99 & 73.68     & 98.67           \\
    \bottomrule
  \end{tabular}
\end{table}

\section{Conclusions}\label{sec:conclusions}
In this work, we scrutinized the omnipresence of prompt tuning in recent continual learning methods in favor of other parameter-efficient fine-tuning (PEFT) methods.
We did this by introducing \propmethod{}, a LoRA-based continual learning method.
We empirically demonstrated that it outperforms its prompt tuning counterpart in domain- and class-incremental learning by a large margin and remains as parameter-efficient.
Furthermore, we improved the unsupervised dataset identification strategy by using the representation of the fine-tuned model.
This change resulted in new state-of-the-art results on Split CIFAR-100.

\bibliographystyle{plain}
\bibliography{references}

\newpage
\begin{appendices}

    \crefalias{section}{appendix}
    
    \section{Related Work}\label{sec:prior_work}
    
Continual learning methods can be broadly classified based on how they retain the information learned in previous datasets. 
\textit{Replay-based} methods tackle catastrophic forgetting by using some additional data which is used when training on the new data\citep{hou2019learning,marra2019incremental,wu2019large,chaudhry2019tiny,buzzega2020dark,prabhu2020gdumb,cha2021co2l}.
These methods store a few data points from previous datasets in a memory of limited size and replay those data points during training.
Memory-free approaches replace true data points with generated or auxiliary data, which is replayed\citep{kemker2018fearnet,ye202learning}.

\textit{Regularization-based} methods oftentimes require no memory and avoid forgetting by adding regularization terms to the loss function.
These terms can either regularize the weights directly to avoid changing important weights\citep{kirkpatrick2017overcoming,ritter2018online} or regularizing activation outputs\citep{li2017learning,dhar2019learning}.

With the advent of large scale pre-trained transformers, memory-free continual learning based on {prompt-tuning}\citep{lester-etal-2021-power} for domain-incremental or class-incremental learning, or adapters\citep{houlsyb2019adapters} for task-incremental learning\citep{ermis2022ada} have been proposed recently.
Learning To Prompt (L2P)\citep{zhou2021learning}, based on prompt-tuning, learns a set of input-dependent prompts that are shared across datasets. Dual-Prompts\citep{wang2022dualprompt} extends
this by learning adding dataset-dependent and dataset-independent prompts at various points in the network.
In addition to this idea, follow-up work proposes to learn components which are combined to prompts at inference time\citep{seale2023codaprompt}.
Works that simplify the problem by learning a per-dataset prompt that are combined for efficient forward transfer exist. 
However, this requires to assume a task-incremental setting where old prompts are not further updated\citep{razdaibiedina2023progressiveprompts} or access to old data\citep{douillard2021dytox}.
S-Prompts overcomes this problem by training assuming a task-incremental setting and then solving the task identification problem at inference time using clustering\citep{wang2022sprompts}.
The work discussed here for continual learning for transformers relies on variations of prompt-tuning or prefix-tuning\citep{li-liang-2021-prefix}. Additionally, S-Prompts is primarily 
shown to work for domain-incremental scenarios. Our method, \propmethod, extends this line of work by using LoRA modules, retains the simplicity of S-Prompts, and is effective at both 
domain incremental and task incremental learning scenarios.

    \section{Vision Transformer}\label{app:sec:vit}
    In this section, we describe the Vision Transformer\citep{dosovitskiy2020vit} (ViT) that we use in this paper. ViT ingests an image $I\in \mathbb{R}^{W\times H\times 3}$, and first extracts patches of size $P\times P$, totalling $\frac{W\times H}{P^2}$ patches per image. Each of these patches is flattened and embedded into a $D$ dimensional space. To this a learned position encoding ($E_{\text{pos}}$) is added, and a special token called the classification ([CLS]) token is concatenated. We refer to this as $X_0 \in \mathbb{R}^{N\times D}$ where $N = \frac{W\times H}{P^2} + 1$. This operation can be represented as
    
    \begin{equation}
        X_0 = [\text{[CLS]}; I_p^1E; \cdots I_p^{(N-1)}E] + E_{\text{pos}}.
    \end{equation}
    This feature representation is processed through $L$ layers of multi-head self attention layers.
     $$\left.
        \begin{array}{ll}
        X^a_l &= \text{MHSA}(X_{l-1}) + X_{l-1} \\ 
        X_l &= \text{FFN}( X^a_l) +  X^a_l
    \end{array}
    \right \} \forall l = 1 \dots L
    $$
    
    The function MHSA consists of mutiple SA modules that function in parallel. Each SA module can be written as
    \begin{equation}
        \text{SA}(X_l) = \mathrm{softmax}\left(\frac{X_lW^l_Q{W^l_K}^TX_l^T}{2\sqrt{d}}\right)X_l\W_V
    \end{equation}
    and the FFN as 
    \begin{equation}
        \text{FFN}(X_l) = \text{GeLU}(W^l_2\text{GeLU}(W^l_1X_l+b^l_1)+b^l_2).
    \end{equation}
    The [CLS] token at $X_L$ is fed into a linear layer $\mathbb{R}^{D} \rightarrow  \mathbb{R}^{C}$ that outputs the logits for classification.
    The set of trainable parameters for fine-tuning is $\{W^l_*, b^l_*\}_{l=1}^L$. 

    \section{Training hyperparameters}\label{app:sec:hps}
    We closely follow the protocol by earlier work to allow for fair comparison~\cite{wang2022sprompts}.
    We adopt their data augmentation which consists of simple horizontal flips and random crops.
    We use a batch size of $128$ and a weight decay of $0.0002$.
    We set learning rates and epochs to minimize training budget.
    In most cases, we use $50$ epochs with the exception of CORe50 where we use $20$.
    As a default, we use a learning rate of $10^{-3}$.
    For CIFAR-100, we use $0.01$, for CORe50, $0.02$.
    Cosine annealing is used to decay the learning rate over time.
    Unless otherwise stated, we use a LoRA rank of $64$.
    We set the number of clusters to $k=5$ as recommended for S-Prompts~\cite{wang2022sprompts} in DIL.
    For CIL, we set the number of clusters to two times the number of new classes, \ie 20 for Split CIFAR-100.
    The choice of number of clusters and the rank is ablated in \cref{sec:experiments,app:sec:ablations}.

\section{Results}
\label{app:results}

In this section, we extend the results in \cref{fig:cil,fig:dil} by comparing \propmethod{} to replay-based methods in \cref{tab:cil,tab:dil}.

For the domain incremental scenario presented in \cref{tab:dil}, we observe that \propmethod{} outperforms replay method with limited buffer sizes on most datasets.
On DomainNet, performance of \propmethod{} is only matched by that of DyTox which uses a replay buffer. %

In \cref{tab:cil}, we present detailed results for Split CIFAR-100.
For fine-tuning, we fine-tune the entire ViT model and mask the outputs for classes not present in an update by setting those logits
to $-\infty$. 
We find that this is important for L2P, without which its performance suffers drastically.
Using ``class-masking'', fine-tuning results in \cref{tab:cil} are substantially higher than 
the ones reported in literature as FT-seq and FT-seq-frozen.
Furthermore, we report the results obtained when training the ViT on all data using LoRA, and fine-tuning the entire model as the upper bound.

\begin{table}[H]
\caption{
Average accuracy results on three domain-incremental benchmarks.
\propmethod{} consistently outperforms alternative approaches even if these have access to previous data.
This includes the self-reported upper bound for S-Prompts which has access to all data.
Results marked %
with ${\dagger}$ from \cite{wang2021learning}, and with ${\ddagger}$ from \cite{wang2022sprompts}.
} 
\label{tab:dil}
\begin{center}
\begin{tabular}{lcHcc}
\toprule 
 Method & Buffer Size & CDDB & CORe50 & DomainNet\\
\midrule
 S-Prompts (upper bound) & \multirow{2}{*}{$\infty$}& 85.50$^{\ddagger}$ & 84.01$^{\ddagger}$ & 63.22$^{\ddagger}$ \\
 LoRA ($r=64$) && 88.75\scriptsize{$\pm$0.10} & 96.15\scriptsize{$\pm$0.07} & 73.62\scriptsize{$\pm$0.02} \\
\midrule
 DyTox~\cite{douillard2021dytox} &  \multirow{7}{*}{50/class} & \textbf{86.21}$^{\ddagger}$ & 79.21$^{\ddagger}$\scriptsize{$\pm$0.10} & 62.94$^{\ddagger}$ \\
 ER~\cite{chaudhry2019tiny} & &- & 80.10$^{\dagger}$\scriptsize{$\pm$0.56} & - \\
 GDumb~\cite{prabhu2020gdumb} &&-& 74.92$^{\dagger}$\scriptsize{$\pm$0.25} & - \\
 BiC~\cite{wu2019large} &&-& 79.28$^{\dagger}$\scriptsize{$\pm$0.30} & - \\
 DER++~\cite{buzzega2020dark} &&-& 79.70$^{\dagger}$\scriptsize{$\pm$0.44} &  -\\
 Co$^2$L~\cite{cha2021co2l} &&-& 79.75$^{\dagger}$\scriptsize{$\pm$0.84} & - \\
 L2P~\cite{wang2021learning} &&-& 81.07$^{\dagger}$\scriptsize{$\pm$0.13} & -\\
\midrule
 EWC~\cite{kirkpatrick2017overcoming} & \multirow{8}{*}{0} & 50.59$^{\ddagger}$ & 74.82$^{\dagger}$\scriptsize{$\pm$0.60} & 47.62$^{\ddagger}$ \\ 
 LwF~\cite{li2017learning} &  & 60.94$^{\ddagger}$ & 75.45$^{\dagger}$\scriptsize{$\pm$0.40} & 49.19$^{\ddagger}$ \\ 
 L2P~\cite{wang2021learning} & & 61.28$^{\ddagger}$ & 78.33$^{\dagger}$\scriptsize{$\pm$0.06} & 40.15$^{\ddagger}$ \\
 S-Prompts~\cite{wang2022sprompts} ($k=5$) & & 74.51$^{\ddagger}$ & 83.13$^{\ddagger}$\scriptsize{$\pm$0.51} & 50.62$^{\ddagger}$ \\
 \propmethod{} ($r=1,\ k=5$) &&80.56\scriptsize{$\pm$0.26}& 84.88\scriptsize{$\pm$0.10} & 67.71\scriptsize{$\pm$0.08}\\
 \propmethod{} ($r=8,\ k=5$) &&81.61\scriptsize{$\pm$0.38}& 85.72\scriptsize{$\pm$0.48} & 68.87\scriptsize{$\pm$0.04}\\
 \propmethod{} ($r=64,\ k=5$) &&85.00\scriptsize{$\pm$0.27}& 85.52\scriptsize{$\pm$0.42} & 69.67\scriptsize{$\pm$0.04}\\
 \propmethod{}++ ($r=64,\ k=5$) &&85.61\scriptsize{$\pm$0.34}& \textbf{86.75}\scriptsize{$\pm$0.40} & \textbf{70.06}\scriptsize{$\pm$0.05}\\
\bottomrule
\end{tabular}
\end{center}
\end{table}

\begin{table}[H]
\small
\caption{
Class-incremental learning on CIFAR-100. \propmethod{} outperforms S-Prompts and \propmethod{}++ all other continual learning methods.
This includes the self-reported upper bound for L2P.
Results marked with * are taken from \cite{wang2021learning}.
We report new results for training on all data using LoRA and full fine-tuning.
}
\label{tab:cil}
\begin{center}
\begin{tabular}{lccc}
\toprule 
 \multirow{2}{*}{\textbf{Method}} & \multirow{2}{*}{\textbf{Buffer size}} & \multicolumn{2}{c}{\textbf{Split CIFAR-100}} \\
& &  Average Acc ($\uparrow$) & Forgetting ($\downarrow$)  \\
\midrule
 L2P (upper bound) & \multirow{3}{*}{$\infty$}& 90.85*\scriptsize{$\pm$0.12} & N/A \\
 LoRA ($r=64$) && 92.49\scriptsize{$\pm$0.07} & N/A \\
 Fine-Tuning && 92.11\scriptsize{$\pm$0.10} & N/A \\
\midrule
  ER \cite{chaudhry2019tiny} &\multirow{6}{*}{50/class}& 82.53*\scriptsize{$\pm$0.17} & 16.46*\scriptsize{$\pm$0.25}  \\
 GDumb \cite{prabhu2020gdumb} & & 81.67*\scriptsize{$\pm$0.02} & N/A \\
 BiC \cite{wu2019large} & & 81.4*2\scriptsize{$\pm$0.85} & 17.31*\scriptsize{$\pm$1.02}  \\
 DER++ \cite{buzzega2020dark} & & 83.94*\scriptsize{$\pm$0.34} & 14.55*\scriptsize{$\pm$0.73}  \\
 Co$^2$L \cite{cha2021co2l} & & 82.49*\scriptsize{$\pm$0.89} & 17.48*\scriptsize{$\pm$1.80} \\
 L2P-R~\cite{wang2021learning} & & 86.31*\scriptsize{$\pm$0.59} & \textbf{5.83}*\scriptsize{$\pm$0.61} \\
  \midrule
   ER \cite{chaudhry2019tiny} & \multirow{6}{*}{10/class} & 67.87*\scriptsize{$\pm$0.57} & 33.33*\scriptsize{$\pm$1.28}  \\
  GDumb \cite{prabhu2020gdumb} & & 67.14*\scriptsize{$\pm$0.37} & N/A \\
 BiC \cite{wu2019large} & & 66.11*\scriptsize{$\pm$1.76} & 35.24*\scriptsize{$\pm$1.64}  \\
 DER++ \cite{buzzega2020dark} & & 61.06*\scriptsize{$\pm$0.87} & 39.87*\scriptsize{$\pm$0.99}  \\
 Co$^2$L \cite{cha2021co2l} & & 72.15*\scriptsize{$\pm$1.32} & 28.55*\scriptsize{$\pm$1.56}  \\
 L2P-R~\cite{wang2021learning} & & 84.21*\scriptsize{$\pm$0.53} & 7.72*\scriptsize{$\pm$0.77}  \\
 \midrule
  FT-seq-frozen & \multirow{14}{*}{0} & 17.72*\scriptsize{$\pm$0.34} & 59.09*\scriptsize{$\pm$0.25} \\ 
 FT-seq & & 33.61*\scriptsize{$\pm$0.85} & 86.87*\scriptsize{$\pm$0.20}  \\
 FT+class masking & & 67.02\scriptsize{$\pm$4.20} & 24.37\scriptsize{$\pm$3.76}  \\
 EWC \cite{kirkpatrick2017overcoming} & & 47.01*\scriptsize{$\pm$0.29} & 33.27*\scriptsize{$\pm$1.17}  \\
 LwF \cite{li2017learning} & & 60.69*\scriptsize{$\pm$0.63} & 27.77*\scriptsize{$\pm$2.17}  \\
 L2P~\cite{wang2021learning} & & 83.83*\scriptsize{$\pm$0.04} & 7.63*\scriptsize{$\pm$0.30}  \\
 S-Prompts~\cite{wang2022sprompts} ($k=5$) & & 57.17\scriptsize{$\pm$1.57} & 19.56\scriptsize{$\pm$0.86} \\
 S-Prompts~\cite{wang2022sprompts} ($k=10$) & & 65.71\scriptsize{$\pm$1.50} & 14.76\scriptsize{$\pm$0.75} \\
 S-Prompts~\cite{wang2022sprompts} ($k=20$) & & 67.31\scriptsize{$\pm$1.34} & 12.47\scriptsize{$\pm$1.49} \\
 \propmethod{} ($r=64,\ k=5$) & & 59.98\scriptsize{$\pm$0.04} & 18.69\scriptsize{$\pm$0.41} \\
 \propmethod{} ($r=64,\ k=10$) & & 68.51\scriptsize{$\pm$0.23} & 10.65\scriptsize{$\pm$0.04} \\
 \propmethod{} ($r=1,\ k=20$) & & 70.87\scriptsize{$\pm$0.23} & 10.16\scriptsize{$\pm$0.19} \\
 \propmethod{} ($r=8,\ k=20$) & & 71.22\scriptsize{$\pm$0.11} & 10.22\scriptsize{$\pm$0.18} \\
 \propmethod{} ($r=64,\ k=20$) & & 71.42\scriptsize{$\pm$0.24} & 10.27\scriptsize{$\pm$0.39} \\
 \propmethod{}++ ($r=1,\ k=20$) & & 85.27\scriptsize{$\pm$0.24} & 6.55\scriptsize{$\pm$0.46} \\
 \propmethod{}++ ($r=64,\ k=20$) & & \textbf{86.47}\scriptsize{$\pm$0.07} & 6.25\scriptsize{$\pm$0.34} \\
\bottomrule
\end{tabular}
\end{center}
\end{table}

\section{Ablations}\label{app:sec:ablations}

    In this section, we study the effect of the number of clusters $k$ on the average accuracy.
    We vary $k$ by fixing all other hyperparameters to the defaults described in \cref{app:sec:hps}.    %

\begin{figure}[H]
        \centering
        \renewcommand\sffamily{}
        \input{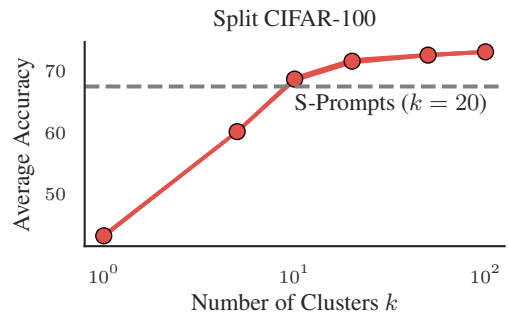}
        \caption{Increasing the number of clusters without changing any other setting. This significantly improves the performance in CIL (Split CIFAR-100) until $k$ equals the number of new classes.}
        \label{fig:ablation-cluster}
    \end{figure}

    In \cref{fig:ablation-cluster}, we observe a similar behavior as that of increasing rank in \cref{fig:ablation-rank} for the number of clusters: more yields better results for CIL, where choosing a large enough number of clusters results in a substantial increase in performance. The advantages of increasing $k$ further diminish very quickly. 
    This is not surprising given that in this scenario, the clusters represent individual classes.
    Therefore, if $k$ is smaller than the number of classes in an update (in this case $10$), the centroids are not able to represent the dataset sufficiently causing dataset detection failures.
    This is clearly demonstrated by the saturation that we achieve once $k$ reaches the number of new classes.
    We find that the choice of $k$ is not too sensitive; above a certain small threshold, its choice has relatively little influence on the results.
    Optimizing it is relatively cheap as it does not require retraining the model.

\end{appendices}

\end{document}